\documentclass[conference,letterpaper]{IEEEtran}
\IEEEoverridecommandlockouts
\usepackage{cite}
\usepackage{amsmath,amssymb,amsfonts}
\usepackage{algorithmic}
\usepackage{graphicx}
\usepackage{textcomp}
\usepackage{xcolor}
\usepackage{svg}
\usepackage{hyperref}
\def\BibTeX{{\rm B\kern-.05em{\sc i\kern-.025em b}\kern-.08em
    T\kern-.1667em\lower.7ex\hbox{E}\kern-.125emX}}
\begin{document}

\title{\LARGE \bf MagNav: A Dual-Core Magnetic Track Guidance Framework for Lighting-Invariant Navigation in Two-Wheeled Robots}

\author{Arpita Kumari*, Anupam Chatterjee and Richa Srivastava
    \thanks{This work was not supported by any external funding.\newline
    *Corresponding author: Arpita Kumari (\href{mailto:arpitaswarnakar30@gmail.com}{arpitaswarnakar30@gmail.com})\newline
    ORCID iDs:\newline
    Arpita Kumari\ 
    \href{https://orcid.org/0009-0007-9219-1084}
    {\includegraphics[height=1.5ex]{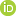}}\ 
    \href{https://orcid.org/0009-0007-9219-1084}
    {0009-0007-9219-1084}\newline
    Anupam Chatterjee\ 
    \href{https://orcid.org/0009-0005-5764-0505}
    {\includegraphics[height=1.5ex]{orcid.png}}\ 
    \href{https://orcid.org/0009-0005-5764-0505}
    {0009-0005-5764-0505}\newline
    Richa Srivastava\ 
    \href{https://orcid.org/0000-0002-8645-3304}
    {\includegraphics[height=1.5ex]{orcid.png}}\ 
    \href{https://orcid.org/0000-0002-8645-3304}
    {0000-0002-8645-3304}\newline
    All authors are with:\newline
    Department of Electronics and Communication Engineering\newline
    Krishna Institute of Engineering \& Technology (KIET),\newline
    Ghaziabad, Delhi-NCR, Uttar Pradesh, India.
    }
}

\maketitle

\begin{abstract}
Two-Wheeled Inverted Pendulum (TWIP) robots are useful for studying how to control systems that are naturally unstable and have fewer actuators than degrees of freedom. Adding autonomous line-following to these robots is challenging because steering and balancing are closely linked. Most existing systems use infrared sensors, which can be affected by changes in lighting, such as sunlight or shadows, making them reliable only indoors. This paper presents a self-balancing robot that can follow a line using a magnetic track guidance system. By using a five-channel analog Hall-effect sensor array, the robot is not affected by optical interference. The control system uses a cascaded PID structure: the inner loop keeps the robot balanced using data from an inertial measurement unit with a complementary filter, while the outer loop adjusts steering based on the magnetic sensor readings. Stepper motors provide precise torque control without needing extra rotary encoders. For comparison, an optical sensor module was also included. Tests show that the magnetic guidance system keeps accurate tracking even in very bright lighting, over 10,000 Lux, while the optical system loses accuracy and sometimes fails. This design provides a reliable, lighting-independent solution for autonomous navigation in places like factories, warehouses, and outdoor paths.
\end{abstract}


\section{Introduction}

The overarching goal in modern autonomous robotics is to achieve reliable navigation for highly manoeuvrable, dynamically stable mobile platforms operating in variable indoor and outdoor environments. Two-wheeled inverted pendulum (TWIP) robots meet this requirement since they have a small footprint and a zero-turning radius, which makes them well-suited for moving through confined spaces and unstructured environments \cite{grasser2002}, \cite{zhang2025}, \cite{hwang2026}, \cite{xin2020}. The fact, though, is that these systems are inherently unstable and underactuated. When one tries to add trajectory tracking abilities to a TWIP system, such as the ability to follow a line autonomously, a great deal of control complexity is introduced since the steering dynamics are directly coupled with the longitudinal balancing loop \cite{jamil2014}, \cite{murcia2016}, \cite{karthika2020}, \cite{tran2021}, \cite{mohsin2022}.

It is generally acknowledged that one of the difficulties in creating autonomous line-following TWIP robots is their dependence on optical sensing. Most traditional systems make use of infrared (IR) reflectance sensors in order to pick up a painted or taped line on the ground. As Pakdaman and others have found in their study \cite{pakdaman2010}, IR-based detection is fundamentally susceptible to interference from surrounding optical sources. Even though more advanced visual-steering methods \cite{li2020} offer additional possibilities, they require substantial computational power. Changes in lighting, such as direct sunlight, shadows, or varying artificial lighting, can make IR sensors less reliable. It is for this reason that the majority of line-following robots are only able to function in carefully controlled indoor environments, which in turn restricts their use in real-world situations.

\begin{figure}[htbp]
\centerline{\includegraphics[width=\columnwidth]{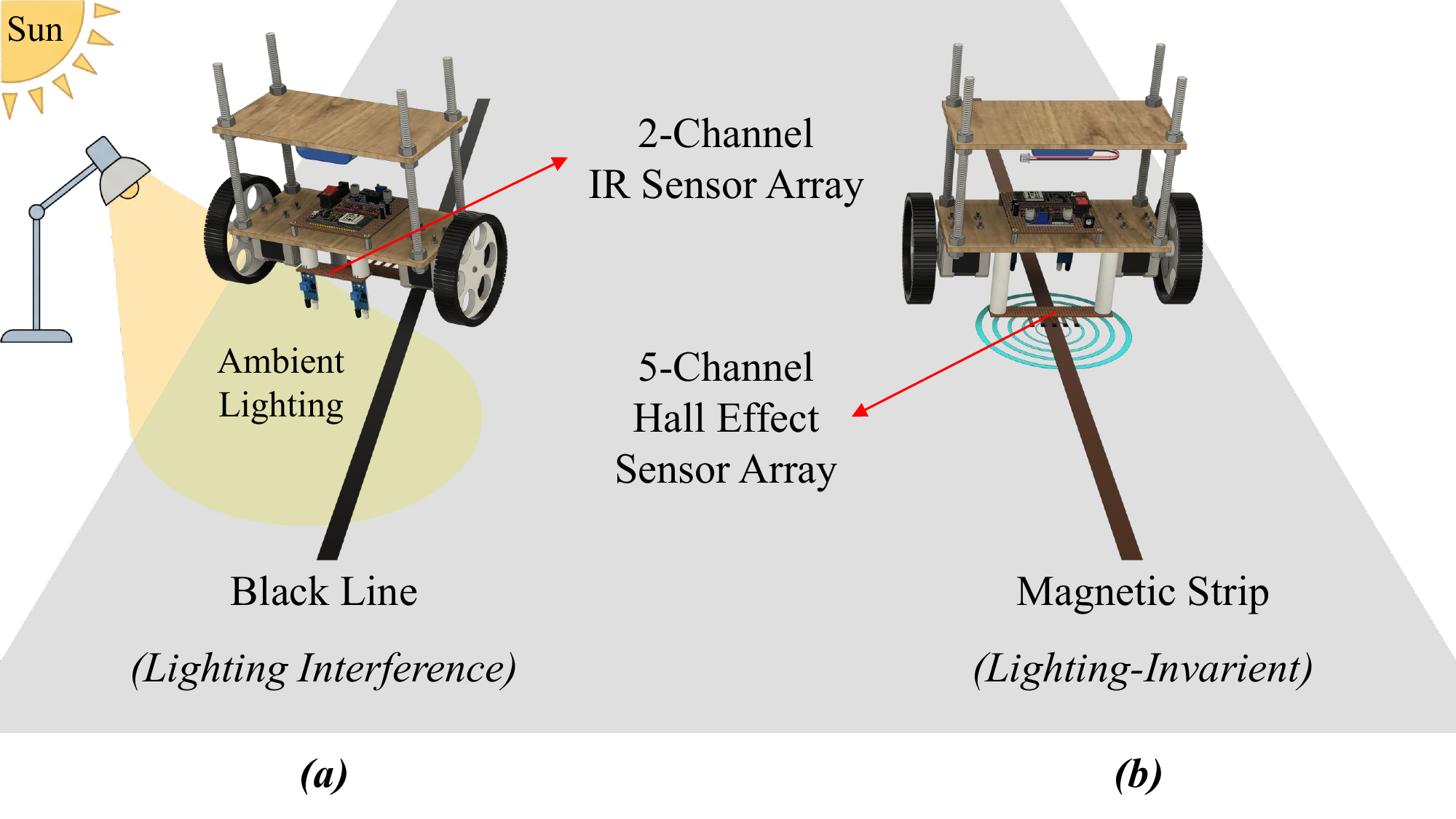}}
\caption{Conceptual comparison of navigation modalities for the Two-Wheeled Self-Balancing Robot. (a) A traditional 2-channel digital IR proximity sensor configuration tracking a black line, which is highly susceptible to ambient lighting interference. (b) The proposed lighting-invariant approach using a custom 5-channel analog Honeywell SS49E Hall-effect sensor array tracking a magnetic strip.}
\label{fig1}
\end{figure}

To overcome these optical limitations, we hypothesise that replacing traditional IR sensing with a magnetic track guidance system will yield a completely lighting-invariant navigation framework. Although magnetic guidance with Hall-effect sensor arrays is a proven technology for highly stable, four-wheeled Automated Guided Vehicles (AGVs) in industrial environments \cite{kamewaka1987}, \cite{lee2012}, its application to the highly non-linear dynamics of a self-balancing mobile manipulator or robot \cite{larimi2013} has been studied only to a limited extent. By integrating a continuous, weighted-average error calculation from a five-channel analog Hall-effect sensor array \cite{ramsden2006} into a cascaded Proportional-Integral-Derivative (PID) control architecture \cite{astrom2004}, \cite{binugroho2015}, the robot can achieve robust steering correction without destabilising the inner high-frequency pitch control loop.

The main conceptual benefit of the system in question is shown in Fig. \ref{fig1}. The robot uses an ESP32 dual-core microcontroller \cite{maier2017} (an improvement over simpler IoT architectures such as those described in \cite{ahmed2021}), which handles both the balancing algorithms and the sensor polling procedures separately to ensure low latency. The pitch is estimated by combining data from the accelerometer and gyroscope of an MPU6050 Inertial Measurement Unit (IMU) using a complementary filter \cite{higgins1975}, \cite{euston2008}. The motors used for actuation are two NEMA 17 stepper motors, which enable precise open-loop rotational control for maintaining equilibrium.

To verify this hypothesis and provide firm proof of concept, the present paper describes a comparative experimental study. Our intention is to show that, under severe ambient lighting conditions (for example, when the intensity of the lighting is extremely high and exceeds 10,000 Lux), the magnetic-guided system exhibits considerably smaller trajectory-tracking error than a similar IR-guided system, the error being measured using the Root Mean Square Error (RMSE) and the Integral Absolute Error (IAE). At the same time, it is necessary for the system to keep the pitch angle variation within acceptable bounds so that the navigation adjustments do not jeopardise dynamic stability. The rest of the paper is organised as follows: the system design is dealt with in Section II, the experimental setup in Section III, the comparative results in Section IV, and the paper concludes with a discussion in Section V.

\section{Methods \& Materials}

We decomposed our system development into three main phases: defining the physical hardware, modelling the system kinematics, and designing the hierarchical control loop to achieve stable balancing and robust trajectory tracking.

\subsection{Physical System Parameters}

\begin{figure}[htbp]
\centerline{\includegraphics[width=\columnwidth]{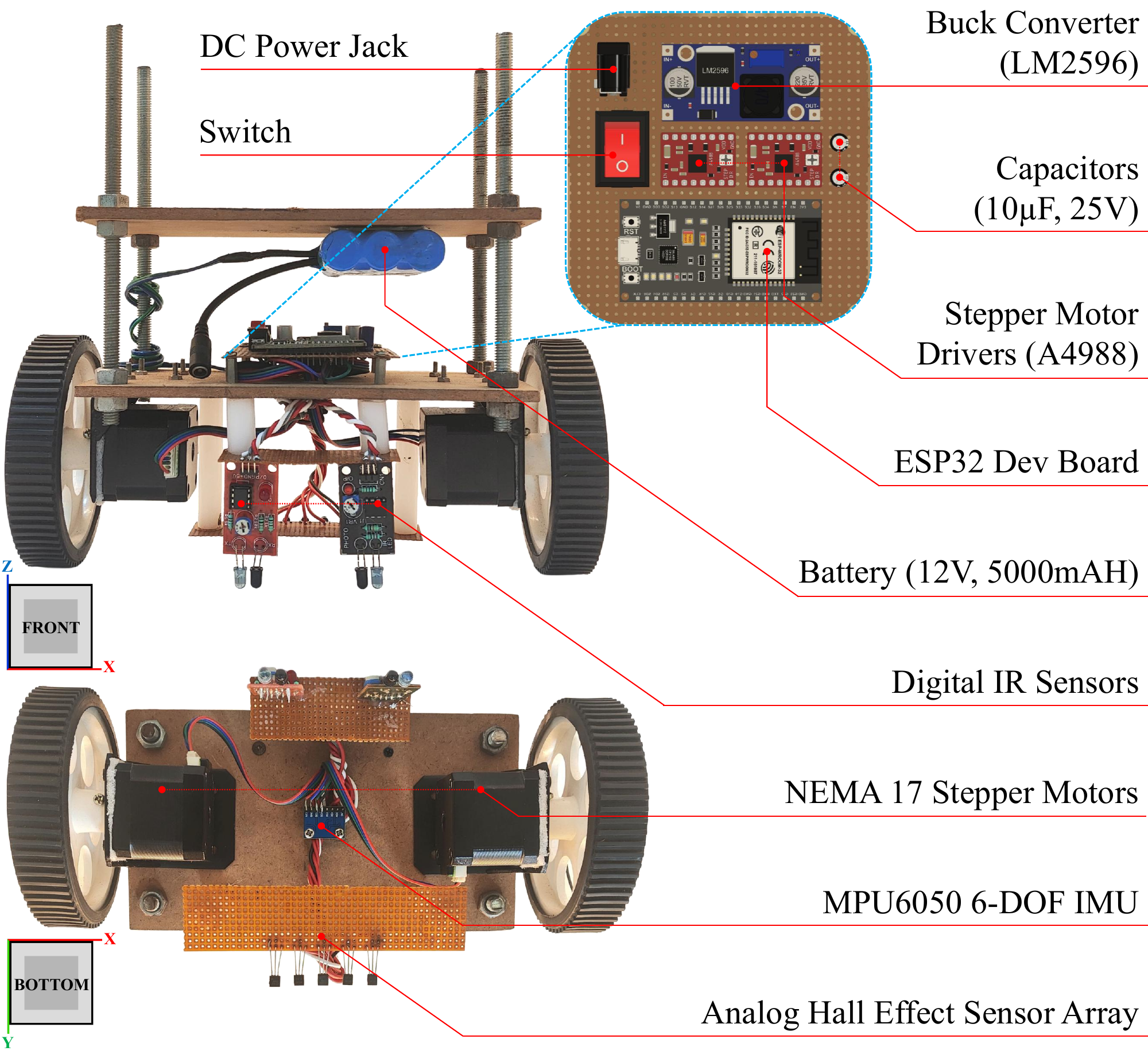}}
\caption{Physical arrangement of the TWIP platform. The upper deck houses the power source, a 12 V, 5000 mAh battery. The lower deck contains the custom-built PCB, which integrates the ESP32 development board, an LM2596 buck converter, and A4988 stepper motor drivers. The bottom view shows the main drive assembly, comprising NEMA 17 stepper motors, an MPU6050 6-DOF IMU for pitch measurement, and two sensor arrays: digital IR and analog Hall-effect sensors.}
\label{fig2}
\end{figure}

The mobile platform was designed with a low centre of mass since this is essential for the passive stability of an inverted pendulum \cite{zhang2025}, \cite{grasser2002}. Figure \ref{fig2} shows the arrangement of the internal components in side, bottom, and top views.

The power is supplied by a 12 V, 5000 mAh battery housed on the upper deck. A custom-built PCB on the lower deck integrates the core ESP32 development board \cite{maier2017}, an LM2596 buck converter for voltage regulation, and A4988 stepper motor drivers, since the ESP32's dual-core design enables the separation of the high-frequency balancing tasks from the lower-priority sensor polling. The actuation is performed using two NEMA 17 stepper motors controlled by the A4988 drivers. Unlike DC motors, stepper motors offer accurate open-loop angular positioning and therefore do not require external rotary encoders in order to estimate wheel odometry. For perception, the robot is equipped with an MPU6050 6-degree-of-freedom (DOF) IMU \cite{invensense2013} mounted at the centre of the chassis. The TWIP platform has a symmetrical structure (it has no intrinsic front or back end), which is why the two sensing systems are located at opposite ends of the chassis, there being a custom five-channel analog Honeywell SS49E Hall-effect sensor array \cite{ramsden2006} on one side and two separate digital IR proximity modules on the other. During testing, the robot moves in the direction of the sensor being tested, so that the same platform can function as both the control setup and the experimental setup without any requirement for reconfiguring the hardware.

\subsection{Kinematic and Dynamic Modelling}

\begin{figure}[htbp]
\centerline{\includegraphics[width=\columnwidth]{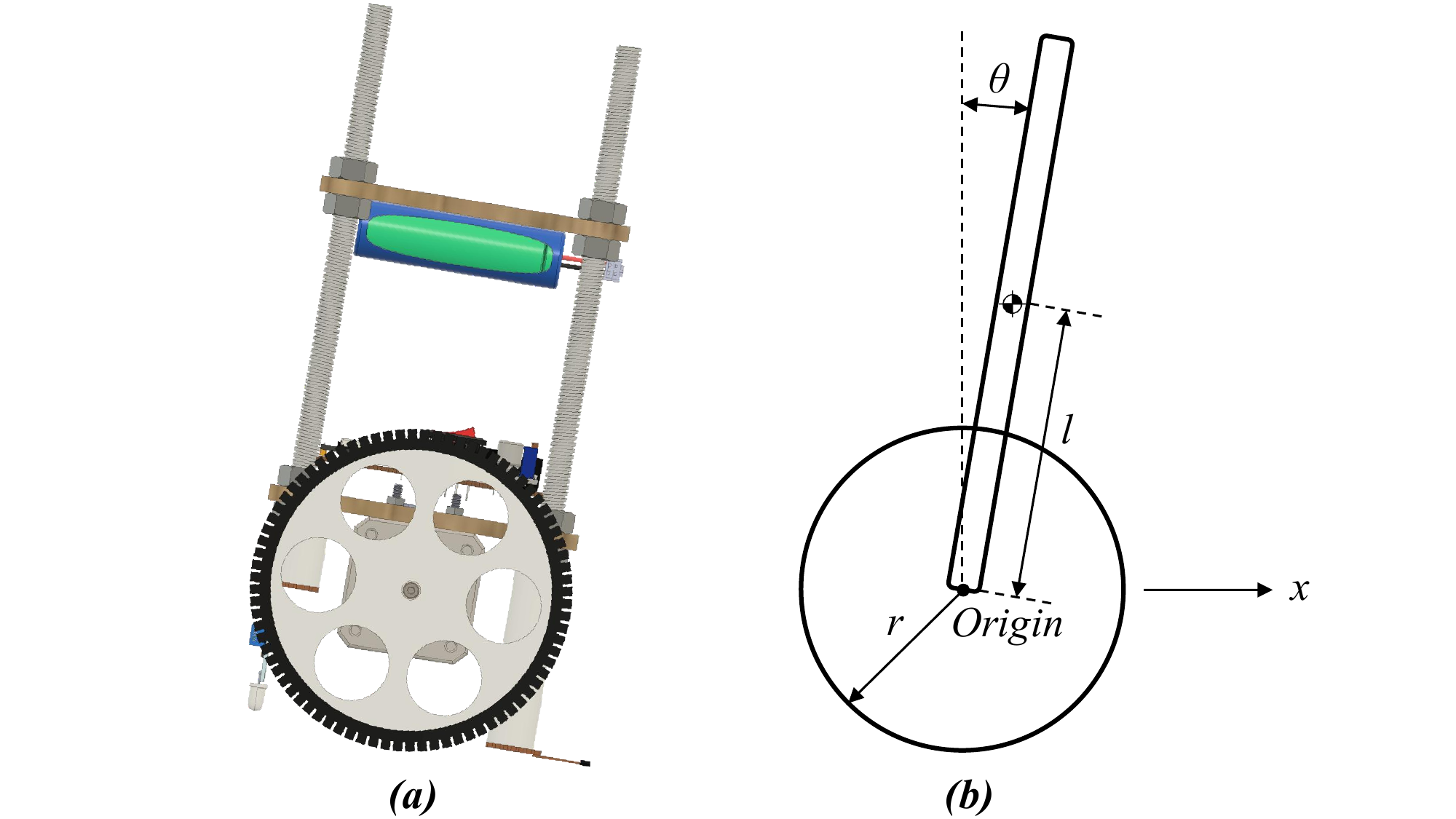}}
\caption{(a) Side CAD view of the physical platform. (b) Free-body diagram of the wheeled inverted pendulum, defining the pitch state ($\theta$), the horizontal translation ($x$), the wheel radius ($r$), and the distance to the centre of mass ($l$).}
\label{fig3}
\end{figure}

The self-balancing robot operates on the principle of a classical two-wheeled inverted pendulum \cite{jamil2014}, \cite{murcia2016}. The primary control objective is to maintain longitudinal pitch stability as the robot manoeuvres horizontally. Fig. \ref{fig3} provides a free-body diagram that maps these parameters to the platform's geometry.

The state of the system is defined by the horizontal position of the cart ($x$), its translational velocity ($\dot{x}$), the pitch angle of the pendulum with respect to the vertical ($\theta$), and its angular velocity ($\dot{\theta}$). Assuming small-angle approximations near the upright equilibrium ($\theta \approx 0$), the dynamics can be expressed in continuous-time state-space form \cite{larimi2013}, \cite{tran2021}. The onboard controller must continuously output corrective wheel accelerations to counteract the gravitational torque acting on the centre of mass located at a distance $l$ from the wheel axle.

\subsection{Hierarchical Control and Sensor Fusion}

\begin{figure}[htbp]
\centerline{\includegraphics[width=\columnwidth]{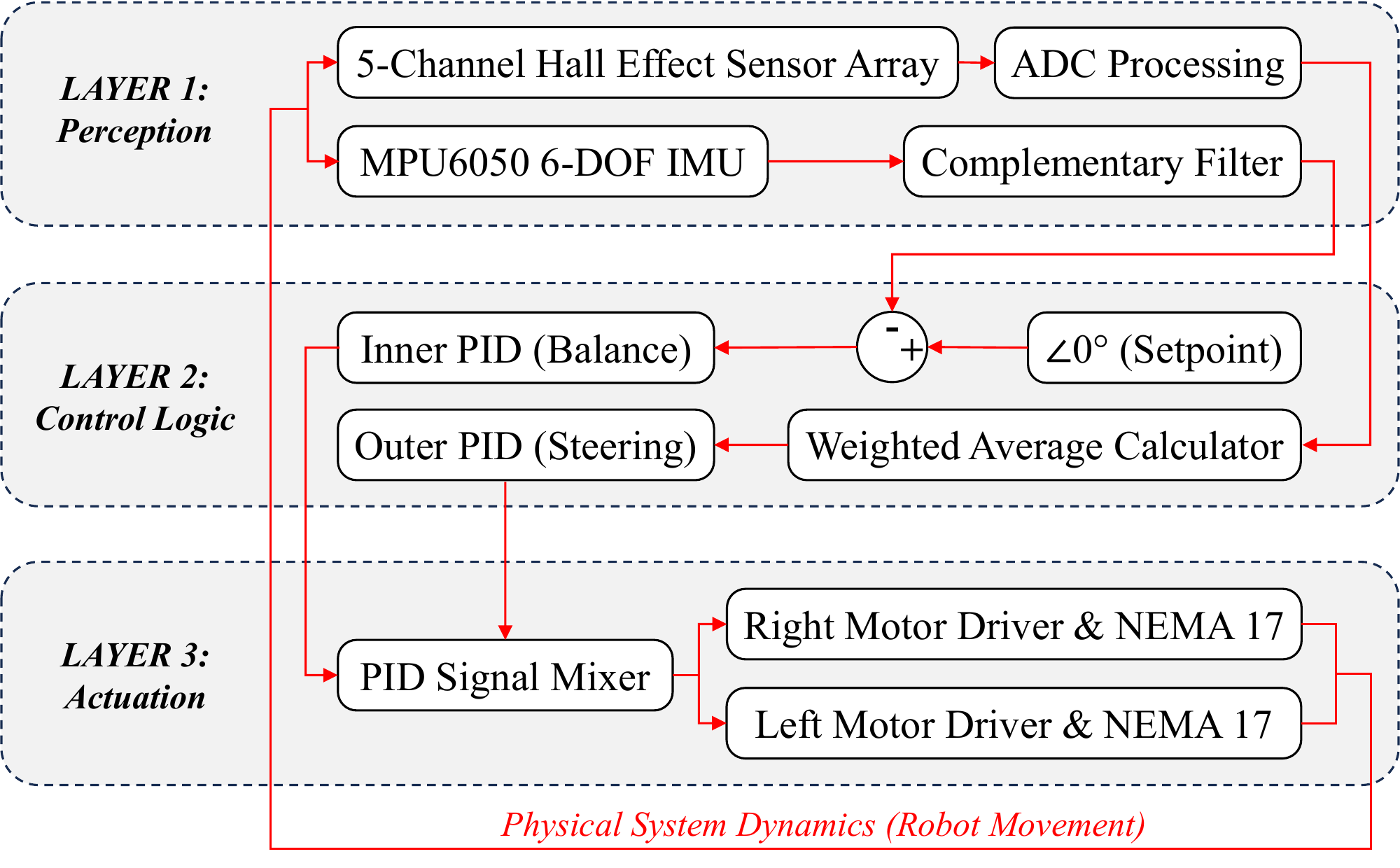}}
\caption{System architecture of the cascaded control framework. The Perception layer fuses IMU and Hall-effect data. The Control Logic layer computes steering and balance corrections, which are combined in the Actuation layer to drive the physical system.}
\label{fig4}
\end{figure}

To manage both unstable balancing dynamics and trajectory tracking, we implemented a cascaded PID control architecture \cite{binugroho2015}, as detailed in the block diagram of Fig. \ref{fig4}. The framework is divided into three distinct layers: Perception, Control Logic, and Actuation.

\subsubsection{Perception Layer}
\begin{figure}[htbp]
\centerline{\includegraphics[width=\columnwidth]{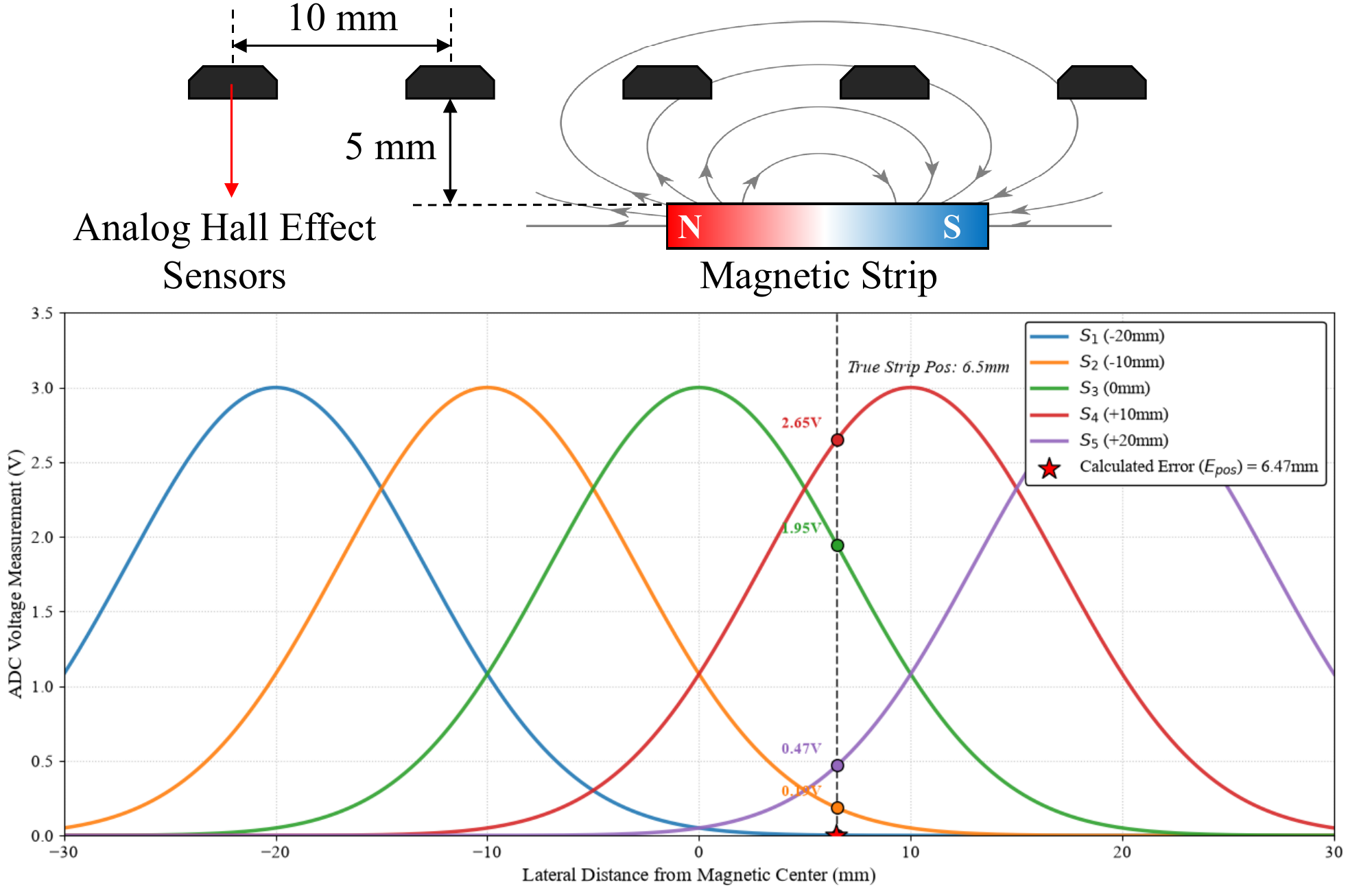}}
\caption{Mathematical conversion of raw analog voltage from the five-channel magnetic sensor array into a high-resolution lateral distance error.}
\label{fig5}
\end{figure}

Accurate estimation of the state is vital for the control loops. The MPU6050 IMU provides raw data from the accelerometer and gyroscope. Yet the raw accelerometer data is highly sensitive to high-frequency vibrational noise from the stepper motors, and the gyroscope is affected by low-frequency drift. In order to overcome these problems, a complementary filter \cite{higgins1975}, \cite{euston2008} is used to combine the two signals, thereby giving a clean, real-time estimate of the pitch angle ($\theta$). At the same time, the five-channel analog Hall-effect sensor array measures the magnetic flux density from the floor strip. The analog-to-digital converter (ADC) is responsible for processing these signals, and in order to determine the continuous lateral position of the magnetic track, a weighted-average algorithm is employed. This interpolation enables a high-precision inter-sensor resolution that is much finer than the actual physical spacing of the sensor array \cite{kamewaka1987}, \cite{lee2012}. As is shown in Fig. \ref{fig5}, the raw analog voltage from the five magnetic sensors is mathematically converted into a single high-resolution distance error.

The error position is computed using the following weighted average formula:
$$E_{pos} = \frac{\sum_{i=1}^{5} (V_i \cdot x_i)}{\sum_{i=1}^{5} V_i}$$
where $E_{pos}$ is the computed continuous lateral error position in millimetres (mm), $V_i$ is the analog voltage reading of the $i$-th sensor in Volts (V), and $x_i$ is the fixed lateral physical distance of that sensor from the centre axis of the robot in millimetres (mm).

\subsubsection{Control Logic Layer}
The control logic is based on a cascaded structure \cite{astrom2004}. The inner PID loop is tasked with maintaining the robot upright. It receives the pitch angle from the complementary filter, compares it with a zero-degree setpoint and calculates the required balancing torque \cite{mohsin2022}. The outer PID loop is for the steering. It takes the lateral line position error from the weighted-average calculator and produces a differential steering command \cite{karthika2020}.

\subsubsection{Actuation Layer}
The signals from the inner and outer PID loops are combined in a PID Signal Mixer, this mixer then determining the final step-frequency pulse trains needed for the left and right wheels, the motor speed being controlled by varying the frequency of the discrete STEP pulses transmitted to each A4988 driver. In the case of forward and backward movement, the balancing output is applied equally to both motors, whereas the steering output is applied differentially (by being added to one wheel and subtracted from the other) in order to produce yaw. The pulse trains are subsequently sent to the A4988 drivers to operate the NEMA 17 stepper motors, thus completing the physical loop.

\section{Experimental Setup}

\begin{figure}[htbp]
\centerline{\includegraphics[width=\columnwidth]{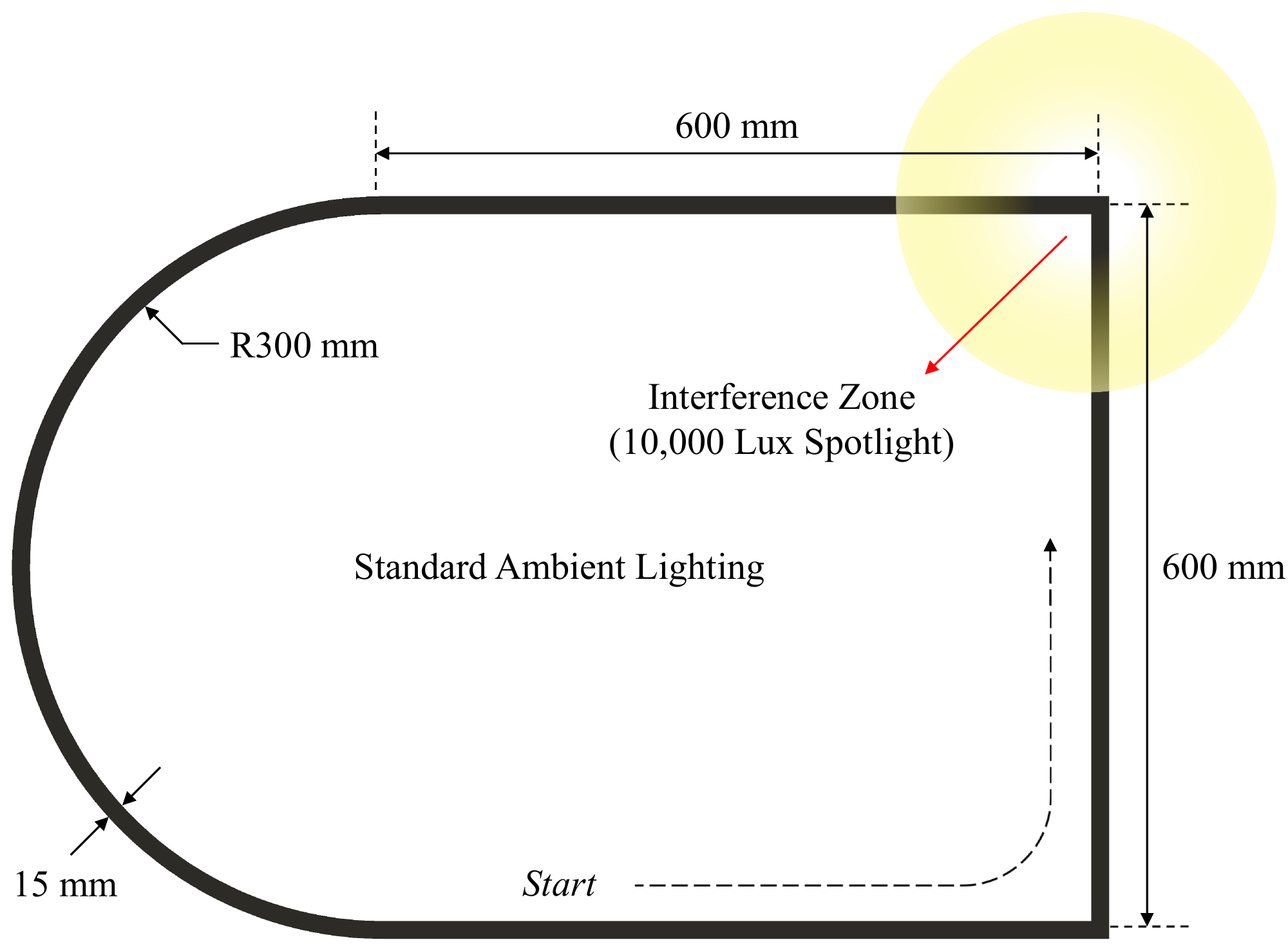}}
\caption{Schematic of the standardised benchmarking track (15 mm track width). The identical geometry was utilised for both optical and magnetic testing phases. An artificial high-intensity interference zone ($>$10,000 Lux) was positioned at the 90-degree vertex to evaluate sensor robustness under extreme ambient lighting.}
\label{fig6}
\end{figure}

The cascaded control architecture was assessed using a standard planar benchmarking track in order to examine its trajectory tracking performance and environmental robustness. As illustrated in Fig. \ref{fig6}, the track is 15 mm wide and has a difficult 90-degree corner. In order to allow for a direct comparative analysis of the two navigation methods, the same track geometry was built two times: once with black IR-absorbent tape for the optical testing phase and again with a continuous magnetic strip for the Hall-effect testing phase.

To put the robot's perception systems to the test, we created particular test situations. Specifically, right at the 90-degree vertex an artificial interference area was introduced which had an illumination intensity greater than 10,000 Lux. This intense lighting condition was deliberately positioned at the point where the greatest steering effort is required in order to closely examine the robustness of the sensors and the stability of the control system under severe optical interference.

\begin{figure}[htbp]
\centerline{\includegraphics[width=\columnwidth]{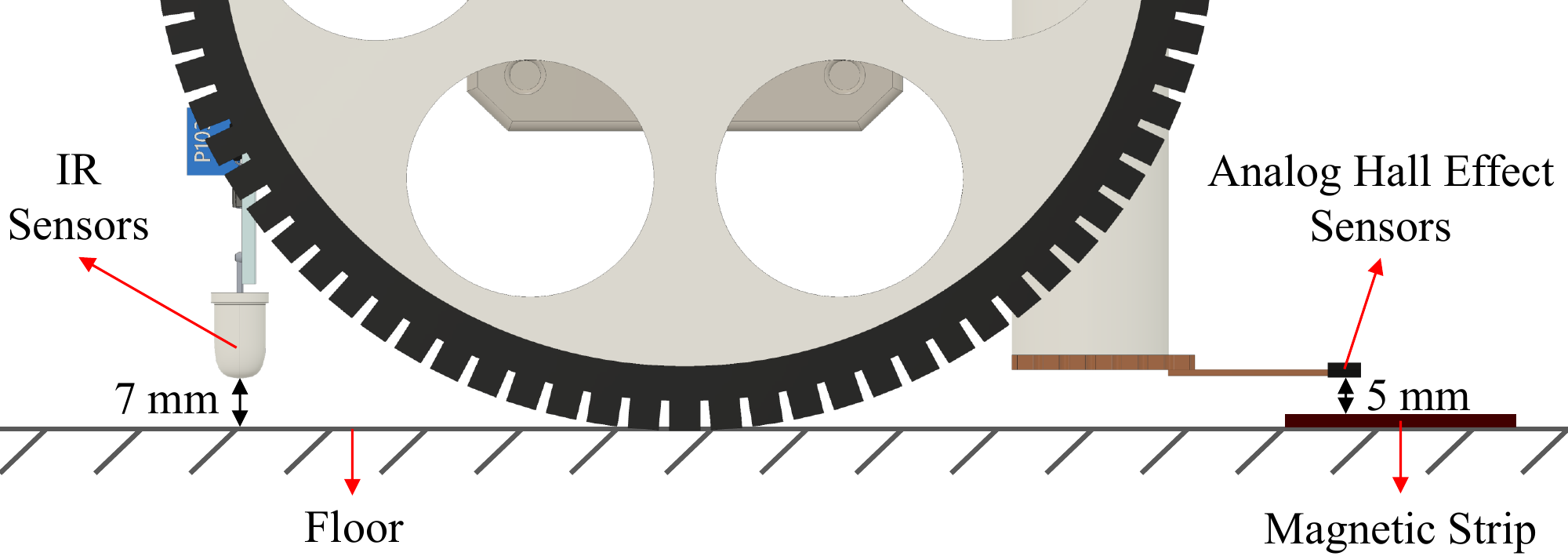}}
\caption{Sensor Mounting \& Clearance Schematic. Both sensor arrays are mounted with a 7.0 mm floor clearance. The 2.0 mm thickness of the magnetic strip results in an effective 5.0 mm sensing air gap for the Hall-effect array, optimising magnetic flux reception while ensuring mechanical safety.}
\label{fig7}
\end{figure}

In order to maintain the integrity of the comparative data, it was essential to fix the physical sensors accurately. As shown in the Sensor Mounting and Clearance Schematic (Fig. \ref{fig7}), each sensor module was firmly attached to the corresponding end of its chassis at a nominal ground clearance of 7.0 mm from the floor \cite{ramsden2006}. However, a critical physical distinction exists between the modalities: while the optical IR modules track flush black tape at a full 7.0 mm clearance, the magnetic guidance relies on a 2.0 mm-thick continuous magnetic strip affixed to the floor. Consequently, the effective air gap for the analog Hall-effect array is reduced to 5.0 mm. This specific configuration optimises the magnetic flux density intersecting the Hall sensors while maintaining sufficient mechanical clearance to prevent dragging during aggressive pitch manoeuvres.

\begin{figure}[htbp]
\centerline{\includegraphics[width=\columnwidth]{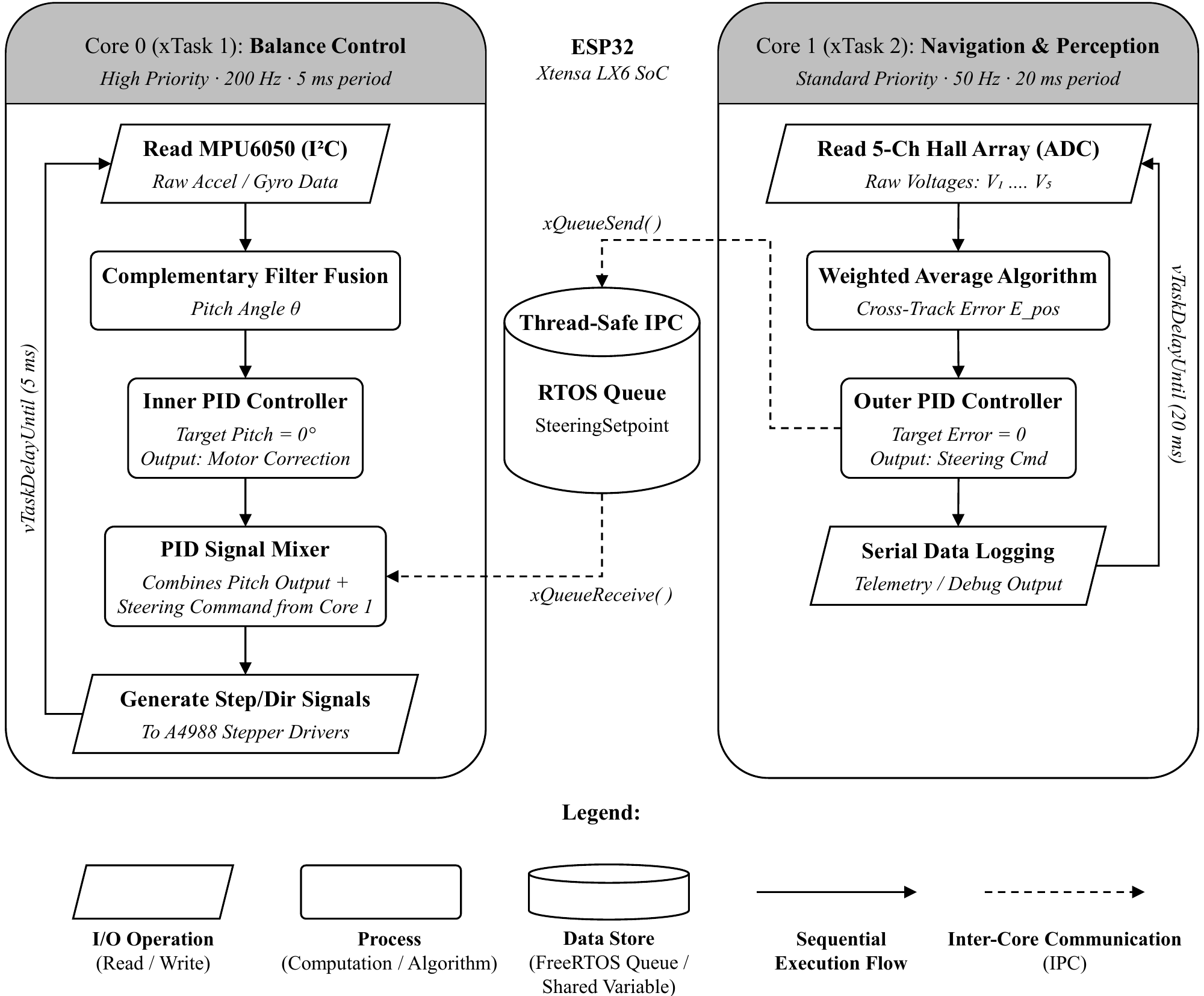}}
\caption{Dual-Core Task Allocation Flowchart mapped to the ESP32 Xtensa LX6 SoC. FreeRTOS separates the deterministic 200 Hz balancing loop on Core 0 from the latency-prone 50 Hz navigation loop on Core 1, utilising thread-safe IPC queues to prevent steering computations from destabilising the platform.}
\label{fig8}
\end{figure}

Moreover, in order to avoid the computationally demanding sensor fusion and path calculation tasks having a negative effect on the basic pitch stability, we made extensive use of the parallel processing features of the ESP32 Xtensa Dual-Core 32-bit LX6 microprocessor \cite{maier2017}. Figure \ref{fig8} shows the Dual-Core Task Allocation Flowchart, which illustrates the strict temporal and spatial separation of the computational tasks controlled by FreeRTOS. Core 0 (\texttt{xTask 1}) was entirely devoted to the high-priority 200 Hz inner balancing loop, carrying out the MPU6050 complementary filter \cite{higgins1975} and driving the stepper motor with timing strictly enforced by \texttt{vTaskDelayUntil}. On the other hand, Core 1 (\texttt{xTask 2}) was responsible for the lower-frequency 50 Hz navigation loop, this loop carrying out the multi-channel ADC polling, computing the weighted-average spatial error, and calculating the outer steering PID. The two separate threads communicate asynchronously through a thread-safe RTOS queue (\texttt{xQueueSend}/\texttt{xQueueReceive}), thus ensuring that the latency caused by heavy analog-to-digital polling cannot interfere with or cause instability in the critical 5 ms inverted pendulum control loop.

\section{Results}

\begin{figure}[htbp]
\centerline{\includegraphics[width=\columnwidth]{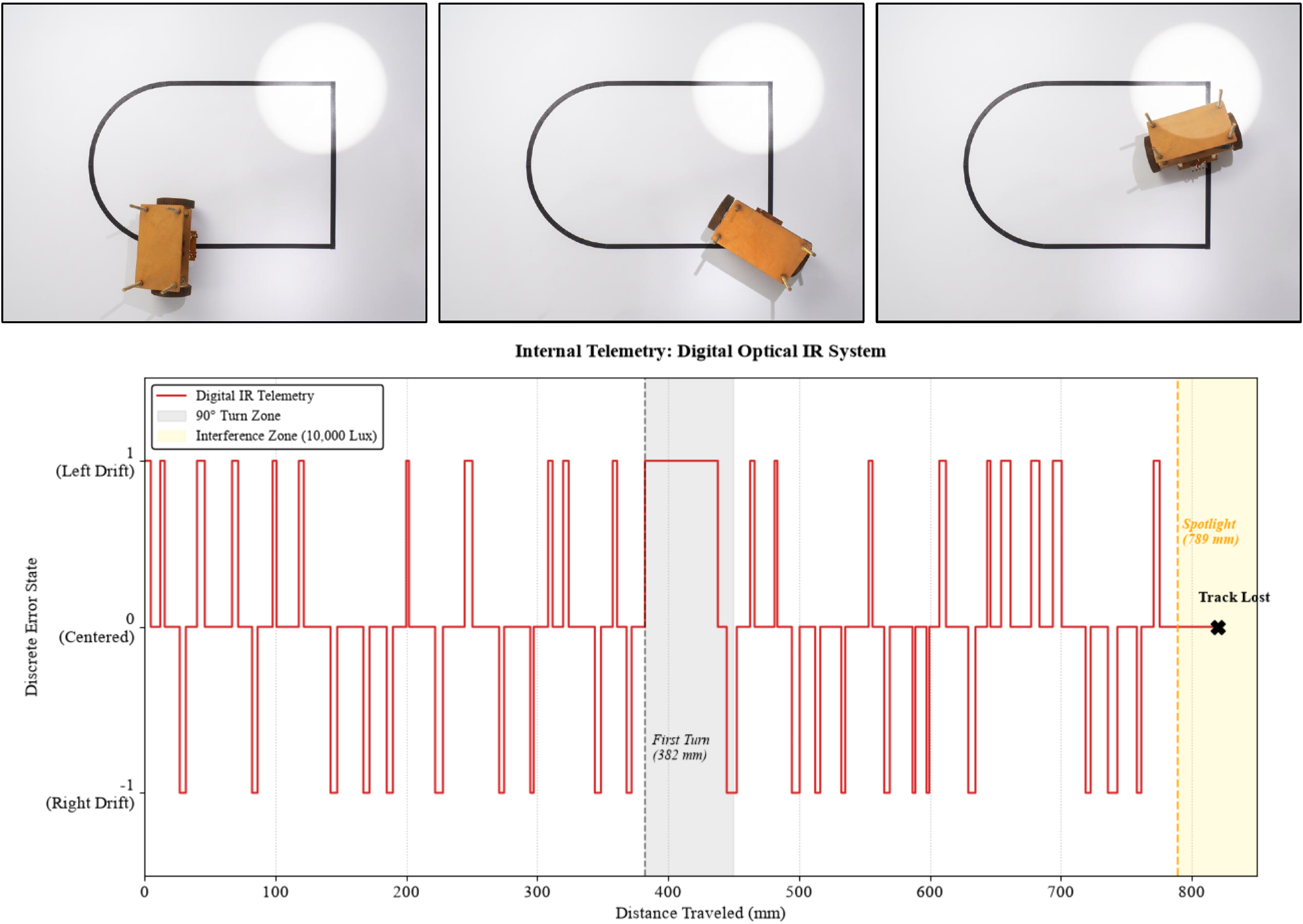}}
\caption{Baseline evaluation: Optical (IR) System Failure. (Top) Physical sequence demonstrating complete path loss upon entering the 10,000 Lux interference zone. (Bottom) Internal telemetry data plotting the Discrete Error State against Distance Travelled, demonstrating the erratic bang-bang control limit cycle and total track loss at 789 mm.}
\label{fig9}
\end{figure}

In order to set up a comparative baseline, we first examined the conventional infrared tracking system. Figure \ref{fig9} shows the severe vulnerability of the optical method. The top sequence shows the robot's actual path, indicating that it successfully navigates the straight section and the first curve. But when the robot enters the 10,000 Lux spotlight interference area at the 90-degree vertex, it fails entirely and moves off the black tape. The synchronised internal telemetry data (shown as the Discrete Error State against Distance Travelled) confirms this failure, highlighting the aggressive limit-cycle oscillations characteristic of discrete sensing. At first, the system displays normal PID oscillations, then has a controlled overshoot at the first turn ($382$ mm). Most importantly, a sharp and uncontrolled increase in tracking error takes place precisely at the interference zone ($789$ mm), leading to a complete loss of track.

\begin{figure}[htbp]
\centerline{\includegraphics[width=\columnwidth]{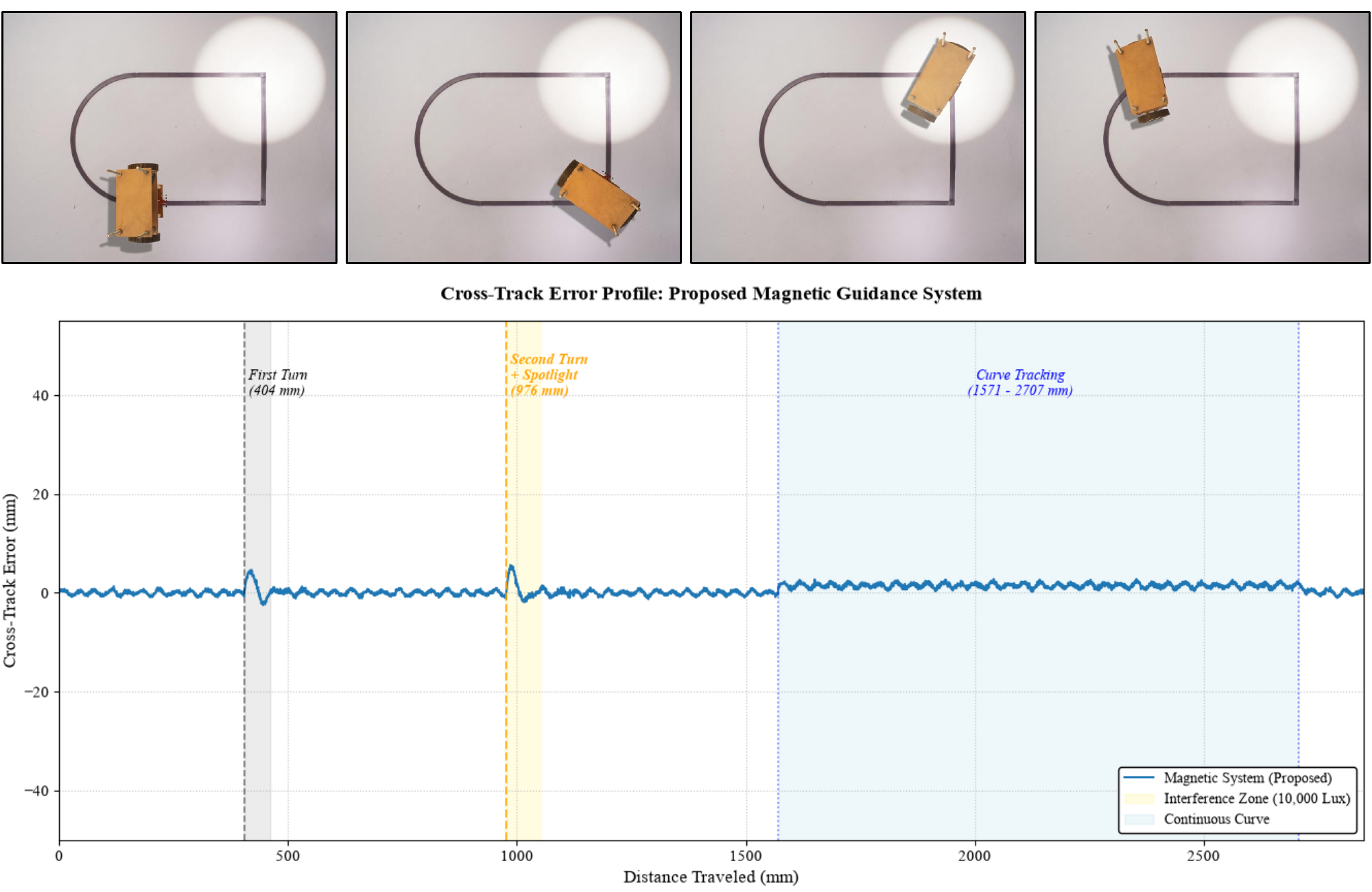}}
\caption{Proposed solution: Magnetic System Success. (At the top) A physical sequence that shows perfect tracking through the $>$10,000 Lux interference area and along the following continuous curve. (At the bottom) Extended telemetry data illustrating complete immunity to optical interference, together with slight, damped overshoots and accurate baseline tracking over a distance of 2742 mm.}
\label{fig10}
\end{figure}

In direct contrast, Fig. \ref{fig10} demonstrates the robust performance of the proposed magnetic guidance system. Utilising the same track geometry and interference conditions, the 5-channel Hall-effect array provided complete immunity to lighting variations. The physical sequence provides conclusive evidence that the robot maintained the 90-degree vertex while under the intense spotlight, then followed a continuous semicircular path. The telemetry data displays the error throughout the entire length of the physical track (from $0$ to $2742$ mm), using the same Y-axis scale in order to allow for direct comparison. The magnetic system shows very precise baseline tracking, having only small and well-damped overshoots at both of the 90-degree turns (at $404$ mm and $976$ mm). Moreover, the data shows a slight and steady offset during the continuous curve tracking section (from $1571$ mm to $2707$ mm), which proves that the stability is maintained.

A quantitative summary of the comparative performance metrics can be found in Table \ref{tab1}. Although Fig. \ref{fig9} shows the IR robot's internal discrete logic telemetry, the maximum physical deviation values for the optical system given in Table \ref{tab1} were independently checked using actual physical measurements. The magnetic system not only completed the full 2742 mm track but also kept considerably tighter error margins throughout all phases, achieving an overall Root Mean Square Error (RMSE) of $2.14$ mm and an Integral Absolute Error (IAE) of $30.9$ mm$\cdot$s. On the other hand, the IR optical system stopped at 789 mm as a result of a catastrophic tracking failure.

\begin{table}[htbp]
\caption{Comparative Performance Metrics}
\label{tab1}
\begin{center}
\scriptsize
\begin{tabular}{|l|c|c|}
\hline
\textbf{Performance Metric} & \textbf{IR Optical (Control)} & \textbf{Magnetic (Proposed)} \\
\hline
Max Error: Straightaway & $\pm 1.8$ mm & $\pm 0.6$ mm \\
Max Error: 90$^{\circ}$ Turn & $16.0$ mm & $8.0$ mm \\
Max Error: Light Zone & Track Lost (Failed) & $8.2$ mm \\
Steady-State Curve Error & N/A (Failed prior) & $1.5$ mm \\
Overall RMSE & N/A (Incomplete run) & $2.14$ mm \\
Overall IAE & N/A (Incomplete run) & $30.9$ mm$\cdot$s \\
Completion Status & Aborted at $789$ mm & Completed ($2742$ mm) \\
\hline
\end{tabular}
\end{center}
\end{table}

Furthermore, the telemetry data obtained from the MPU6050 indicated that the rapid steering corrections required at the 90-degree corners did not disturb the robot's longitudinal dynamic equilibrium. Throughout the entire journey along the 2742 mm magnetic track, the pitch angle standard deviation remained very low ($\sigma < 1.1^\circ$), thus showing that the cascaded PID structure was effective in separating the inner balancing loop from external steering disturbances.

\section{Discussion \& Conclusions}

The primary hypothesis in this study, that magnetic track guidance can provide a lighting-invariant navigation framework for highly unstable, underactuated platforms, has been proven correct. We integrated a 5-channel analog Hall-effect sensor array into a cascaded Proportional-Integral-Derivative (PID) control architecture to eliminate the optical vulnerabilities that plague traditional infrared-guided Two-Wheeled Inverted Pendulum (TWIP) robots. The results of our comparative experimental analysis indicated that although the conventional optical system experienced catastrophic path loss when subjected to severe ambient optical interference ($>$10,000 Lux), the proposed magnetic structure was able to maintain reliable and continuous trajectory tracking. The system achieved this by combining high-frequency longitudinal pitch stabilisation with accurate lateral steering corrections and thus kept control of the platform's complex dynamics without losing equilibrium.

The research has important practical consequences for the robotics industry as a whole, as it now makes it possible to reliably use robots that are inherently unstable yet highly manoeuvrable in unstructured, real-world environments by separating navigation reliability from environmental illumination. This lighting-invariant capability is particularly critical for industrial applications where ambient conditions cannot be strictly controlled. For example, self-balancing mobile manipulators and transport robots are now able to operate smoothly on factory floors with varying artificial lighting, to navigate outdoor routes that are exposed to direct sunlight as well as changing cloud cover, or to move around in warehouses featuring deep and irregular shadows, all situations in which traditional optical line followers would reliably fail.

While the magnetic guidance system demonstrates exceptional robustness to optical interference, we acknowledge certain limitations inherent to the design. Mainly, the system requires the physical installation of a continuous magnetic strip, which makes the infrastructure less flexible and more labour-intensive to alter than virtual routing techniques such as visual SLAM or LiDAR-based mapping \cite{cadena2016}. Future work will focus on overcoming these limitations and increasing the system's capabilities. A promising approach involves combining the high-resolution analog magnetic array with a discrete Radio Frequency Identification (RFID) system in order to achieve absolute spatial localisation and enable complex intersection routing. Additionally, replacing the fixed-gain outer steering loop with an adaptive PID controller \cite{astrom2004} could allow the robot to auto-tune its steering parameters in real-time, further minimising the steady-state error observed during continuous curve navigation.

\section*{Acknowledgment}
AI-assisted tools were used during the preparation of this manuscript to improve language, grammar, readability, and overall structure. We reviewed and revised all AI-assisted content and take full responsibility for the accuracy, originality, and final content of the manuscript.

\bibliographystyle{IEEEtran}
\bibliography{references}

\end{document}